\documentclass[10pt,twocolumn,letterpaper]{article}

\usepackage{iccv}
\usepackage{times}
\usepackage{epsfig}
\usepackage{graphicx}
\usepackage{amsmath}
\usepackage{amssymb}
\usepackage{algorithm}
\usepackage{algorithmic}

\usepackage{rotating}
\usepackage{placeins}
\usepackage{multirow}
\usepackage{slashbox}
\usepackage{bm}
\usepackage{lineno}
\usepackage{color}

\DeclareMathOperator*{\argmax}{arg\,max}
\DeclareMathOperator*{\argmin}{arg\,min}
\DeclareMathOperator*{\tr}{Tr}

\usepackage[pagebackref=true,breaklinks=true,letterpaper=true,colorlinks,bookmarks=false]{hyperref}

\iccvfinalcopy 

\def\iccvPaperID{74} 

\ificcvfinal\fi
\begin{document}

\title{Kernelized Multiview Projection}

\author{Mengyang Yu \qquad Li Liu \qquad Ling Shao\\
Department of Computer Science and Digital Technologies\\
Northumbria University, Newcastle upon Tyne, NE1 8ST, UK\\
{\tt\small m.y.yu@ieee.org, li2.liu@northumbria.ac.uk, ling.shao@ieee.org}
}

\maketitle

\begin{abstract}
Conventional vision algorithms adopt a single type of feature or a simple concatenation of multiple features, which is always represented in a high-dimensional space. In this paper, we propose a novel unsupervised spectral embedding algorithm called Kernelized Multiview Projection (KMP) to better fuse and embed different feature representations. Computing the kernel matrices from different features/views, KMP can encode them with the corresponding weights to achieve a low-dimensional and semantically meaningful subspace where the distribution of each view is sufficiently smooth and discriminative. More crucially, KMP is linear for the reproducing kernel Hilbert space (RKHS) and solves the out-of-sample problem, which allows it to be competent for various practical \nobreak applications. Extensive experiments on three popular image datasets demonstrate the effectiveness of our multiview embedding algorithm.
\end{abstract}

\section{Introduction}

Traditional computer vision techniques are mainly based on single feature representations, either global \cite{DBLP:journals/ijcv/SanchezPMV13} or local \cite{DBLP:conf/cvpr/BoimanSI08}. For local methods, descriptors such as SIFT \cite{DBLP:journals/ijcv/Lowe04} are computed for each detected or densely sampled point, then the Bag-of-Words scheme or its improved version is employed to embed these local features into a whole representation. On the one hand, local feature based methods tend to be more robust and effective in challenging scenarios, while this kind of representation is often not precise and informative because of the quantization error during the codebook construction and the loss of structural relationships among local features. On the other hand, global \nobreak representations \cite{DBLP:journals/ijcv/OlivaT01,dalal2005histograms} describe the image as a whole. Unfortunately, global methods are sensitive to shift, scaling, occlusion and cluttering, which commonly exist in realistic images.

Notwithstanding the remarkable results achieved by both local and global methods in some cases, most of them are still based on a single view (feature representation). In realistic applications, variations in lighting conditions, intra-class differences, complex backgrounds and viewpoint and scale changes all lead to obstacles for robust feature extraction. Naturally, single representations cannot handle realistic tasks to a satisfactory extent.

In practice, a typical sample can be represented by different views/features, e.g., gradient, shape, color, texture and motion. Generally speaking, these views from different feature spaces always maintain their particular statistical characteristics. Accordingly, it is desirable to incorporate these heterogeneous feature descriptors into one compact representation, leading to the multiview learning approaches. These  techniques have been designed for multiview data classification \cite{zien2007multiclass}, clustering \cite{bickel2004multi} and feature selection \cite{zhao2008multi}. For such multiview learning tasks, the feature representations are usually very high-dimensional for each view. However, little effort has been paid to learning low-dimensional and compact representations for multiview computer vision tasks. Thus, how to obtain an effective low-dimensional embedding to discover the discriminative information from all views is a worthy research topic, since the effectiveness and efficiency of the methods drop exponentially as the dimensionality increases, which is commonly referred to as the curse of dimensionality.

Existing multiview embedding techniques include the multiview spectral embedding (MSE) \cite{xia2010multiview} and the multiview stochastic neighbor embedding (m-SNE) \cite{xie2011m}, which have explored the locality information and probability distributions for the fusion of multiview data respectively. Recently, Han et al. \cite{DBLP:journals/tcsv/HanWTSZJ12} proposed a sparse unsupervised dimensionality reduction to obtain a sparse representation for multiview data. However, these methods are only defined on the training data and it remains unclear how to embed the new test data due to their nonlinearity. In other words, they suffer from the \emph{out-of-sample} problem \cite{DBLP:conf/nips/BengioPVDRO03}, which heavily restricts their applicability in realistic and large-scale vision tasks.

In this paper, to tackle the \emph{out-of-sample} problem, we propose a novel unsupervised multiview subspace learning method called kernelized multiview projection (KMP), which can successfully learn the projection to encode different features with different weights achieving a semantically meaningful embedding. KMP considers different probabilistic distributions of data points and the locality information among data simultaneously. Different from the measurement of locality information in the locality preserving projections (LPP) \cite{niyogi2004locality} and the locally linear embedding (LLE) \cite{roweis2000nonlinear}, an $\ell^1$-graph \cite{DBLP:journals/tip/ChengYYFH10,DBLP:conf/icpr/LiuS14} is applied to generate the similarity matrix, which is shown to be more robust to data noise and automatically sparse. Moreover, the $\ell^1$-graph can also adaptively discover the natural neighborhood information for each data point.

Instead of using the multiview features directly, the kernel matrices from multiple views enable KMP to normalize the scales and the dimensions of different features. In fact, we show that the fusion of multiple kernels is actually the concatenation of features in the high-dimensional reproducing kernel Hilbert space (RKHS), while the learning phase of KMP remains in the low-dimensional space. Having obtained kernels for each view in RKHS, KMP can not only fuse the views by exploring the complementary property of different views as multiple kernel learning (MKL) \cite{DBLP:journals/jmlr/LanckrietCBGJ03,DBLP:journals/jmlr/GonenA11,vedaldi2009multiple}, but also find a common low-dimensional subspace where the distribution of each view is sufficiently smooth and discriminative. Note that multiview learning techniques are used to fuse different views/features while MKL is used to combine different kernel functions.

The rest of this paper is organized as follows. In Section \ref{rw}, we give a brief review of the related work. The details of our method are described in Section \ref{method}. Section \ref{exp} reports the experimental results. Finally, we conclude this paper in Section \ref{con}.

\section{Related Work}\label{rw}

A simple multiview embedding framework is to concatenate the feature vectors from different views together as a new representation and utilize an existing dimensionality reduction method directly on the concatenated vector to obtain the final multiview representation. Nonetheless, this kind of concatenation is not physically meaningful because each view has a specific characteristic. And, the relationship between different views is ignored and the complementary nature of intrinsic data structure of different views is not sufficiently explored.

One feasible solution is proposed in \cite{long2008general}, namely, distributed spectral embedding (DSE). For DSE, a spectral embedding scheme is first performed on each view, respectively, producing the individual low-dimensional representations. After that, a common compact embedding is finally learned to guarantee that it would be similar with all single-view's representations as much as possible. Although the spectral structure of each view can be effectively considered for learning a multiview embedding via DSE, the complementarity between different views is still neglected.

To effectively and efficiently learn the complementary nature of different views, multiview spectral embedding (MSE) is introduced in \cite{xia2010multiview}. The main advantage of MSE is that it can simultaneously learn a low-dimensional embedding over all views rather than separate learning as in DSE. Additionally, MSE shows better effectiveness in fusing different views in the learning phase.

However, both DSE and MSE are based on nonlinear embedding, which leads to a serious computational complexity problem and the \emph{out-of-sample} problem \cite{DBLP:conf/nips/BengioPVDRO03}. In particular, when we apply them to classification or retrieval tasks, the methods have to be re-trained for learning the low-dimensional embedding when new test data are used. Due to their nonlinearity nature, this will cause heavily computational costs and even become impractical for realistic and large-scale scenarios.

Towards solving the \emph{out-of-sample} problem for multiview embedding, we propose a unsupervised projection method, namely, KMP. It is noteworthy that, as a linear method, a projection is learned via the proposed KMP using all of the training data. Nevertheless, different from non-linear approaches, once the learning phase finishes, the projection will be fixed and can be directly applied to embed any new test sample without re-training.


\section{Kernelized Multiview Projection}\label{method}

\subsection{Notations}
Given $N$ training samples $\{S_1, \cdots, S_N\}$ and $M$ different descriptors for multiview feature extraction, $X^i_p \in \mathbb{R}^{D_i}$ represents the feature vector for the $i$-th view and $p$-th sample. Since the dimensions of various descriptors are different, kernel matrices $K_1, \cdots, K_M \in \mathbb{R}^{N \times N}$ are constructed by the kernel functions such as the RBF kernel and the polynomial kernel, for the fusion of different views in the same scale. Our task is to output an optimal projection matrix $P \in \mathbb{R}^{N \times d}$ and weights $(\alpha_1, \cdots, \alpha_M)$ satisfying $\sum^M_{i=1} \alpha_i =1$ for kernel matrices such that the fused feature matrix $Y = [\mathbf{y}_1, \cdots, \mathbf{y}_N]^T = KP = (\sum^{M}_{i=1} \alpha_i K_i)P$ can represent original multiview data comprehensively.


\subsection{Formulation of KMP}
The projection learning of KMP is based on the similarity matrix $W_i$ for the $i$-th view, $i=1,2, \cdots, M$. For each view, we value the similarity of each sample pair by using the neighbors of each point. The construction of $W_i$ is illustrated below via the $\ell^1$-graph \cite{DBLP:journals/tip/ChengYYFH10}, which is demonstrated to be robust to data noise, automatically sparse and adaptive to the neighborhood.

\paragraph{Similarity construction} For each $X^i_p$, we find the coefficients $\bm{\beta} \in \mathbb{R}^{N -1}$ such that $X^i_p = B \bm{\beta}$, where $B = [X^i_1, \cdots, X^i_{p-1}, X^i_{p+1}, \cdots, X^i_N] \in \mathbb{R}^{D_i \times (N-1)}$. Considering the noise effect, we can rewrite it as $X^i_p = B' \bm{\beta}'$, where $B' = [B, I] \in \mathbb{R}^{D_i \times (D_i + N -1)}$ and $\bm{\beta}' \in \mathbb{R}^{D_i + N -1}$. Thus, seeking the sparse representation for $X^i_p$ leads to the following optimization problem:
\begin{equation}\label{l1graph}
  \argmin_{\bm{\beta}'} \|X^i_p - B' \bm{\beta}'\|_2 , ~\text{s.t.}~ \|\bm{\beta}'\|_1 < \varepsilon,
\end{equation}
where $\varepsilon$ is the parameter with a small value. This problem can be solved by the orthogonal matching pursuit \cite{pati1993orthogonal}.

Considering different probabilistic distributions that exist over the data points and the natural locality information of the data, we first employ the Gaussian mixture model (GMM) on the training data for each view. On the one hand, it has been proved that data in the high-dimensional space do not always follow the same distribution, but are naturally clustered into several groups. On the other hand, realistic data distributions basically follow the same form, i.e., Gaussian distribution. In this case, $G$ clusters are obtained by the unsupervised GMM clustering for each view. Thus, we can solve the above problem (\ref{l1graph}) using the data from the same cluster to represent each point rather than the whole data points $B$, which is also regarded as a solution to alleviate the computational complexity of problem (\ref{l1graph}).

In particular, for $\bm{\beta}' = (\beta_1, \cdots, \beta_{D_i + N -1})$, we can first set $\beta_q = 0$ if $X^i_q$ and $X^i_p$ are in different clusters, $\forall q \neq p$, then solve the above problem. Now the similarity matrix $W_i \in \mathbb{R}^{N \times N}$ can be defined as: $(W_i)_{pp} = 0$, $\forall p$, $(W_i)_{pq} = |\beta_q|$ if $q < p$, and $(W_i)_{pq} = |\beta_{q-1}|$ if $q > p$. To ensure the symmetry, we update $W_i \leftarrow (W_i^T + W_i)/2$. Then we set the diagonal matrix $D_i \in \mathbb{R}^{N \times N}$ with $(D_i)_{pp} = \sum_q (W_i)_{pq}$ and the Laplacian matrix $L_i = D_i - W_i$ for each view $i$.

\paragraph{Multiview kernel fusion} Due to the complementary nature of different descriptors, we assign different weights for different views. The goal of KMP is to find the basis of a subspace in which the lower-dimensional representation can preserve the intrinsic structure of original data. Therefore, we impose a set of nonnegative weights $\alpha= (\alpha_1, \cdots, \alpha_M)$ on the similarity matrices $W_1, \cdots, W_M$ and we have the fused similarity matrix $W = \sum^M_{i=1} \alpha_i W_i$, fused diagonal matrix $D = \sum^M_{i=1} \alpha_i D_i$ and the fused Laplacian matrix $L = \sum^M_{i=1} \alpha_i L_i$.

For the kernel matrix, we also define the fused kernel matrix $K= \sum^M_{i=1} \alpha_i K_i$. In fact, suppose $\phi_i$ is the substantial feature map for kernel $K_i$, i.e., $K_i = \phi_i(X^i)^T \phi_i(X^i)$, then the fused kernel value is computed by the feature vector concatenated by the mapped vectors via $\phi_1, \cdots, \phi_M$, since we have
\begin{equation*}
\begin{split}
  K & = \sum^M_{i=1} \alpha_i K_i  = \sum^M_{i=1} \alpha_i \phi_i(X^i)^T \phi_i(X^i) \\
    & = \left[
         \begin{array}{c}
           \sqrt{\alpha_1} \phi_1(X^1) \\
           \vdots \\
           \sqrt{\alpha_M} \phi_M(X^M) \\
         \end{array}
       \right]^T
       \left[
         \begin{array}{c}
           \sqrt{\alpha_1} \phi_1(X^1) \\
           \vdots \\
           \sqrt{\alpha_M} \phi_M(X^M) \\
         \end{array}
       \right]\\
    & = \phi(X)^T \phi(X),
\end{split}
\end{equation*}
where $\phi(\cdot) = [\sqrt{\alpha_1} \phi_1(\cdot)^T, \cdots,  \sqrt{\alpha_M} \phi_M(\cdot)^T]^T$ is the fused feature map and $X = (X^1, \cdots, X^M)$ is the $M$-tuple consisting of features from all the views.

To preserve the fused locality information, we need to find the optimal projection for the following optimization problem:
\begin{equation}\label{}
  \argmin_{\mathbf{v}} \sum_{p,q} \|\mathbf{v}^T \psi_p - \mathbf{v}^T \psi_q\|_2^2 (W)_{pq},
\end{equation}
where $\psi_p$ is the fused mapped feature, i.e., $[\psi_1, \cdots, \psi_N] = \phi(X)$. Through simple algebra derivation, the above optimization problem can be transformed to the following form:
\begin{equation}\label{eqlpp1}
  \argmin_{\mathbf{v}} \tr(\mathbf{v}^T \phi(X) L \phi(X)^T \mathbf{v}).
\end{equation}
With the constraint $\tr(\mathbf{v}^T \phi(X) D \phi(X)^T \mathbf{v}) = 1$, minimizing the objective function in Eq. (\ref{eqlpp1}) is to solve the following generalized eigenvalue problem:
\begin{equation}\label{eiglpp}
  \phi(X) L \phi(X)^T \mathbf{v} = \lambda \phi(X) D \phi(X)^T \mathbf{v}.
\end{equation}
Note that each solution of problem (\ref{eiglpp}) is a linear combination of $\psi_1, \cdots, \psi_N$, and there exists an $N$-tuple $\mathbf{p} = (p_1, \cdots, p_N)^T \in \mathbb{R}^N$ such that $\mathbf{v} = \sum^N_{i=1} p_i \psi_i = \phi(X) \mathbf{p}$. For matrix $V$ consisting of all the linearly independent solutions of problem (\ref{eiglpp}), there exists a matrix $P$ such that $V= \phi(X)P$. Therefore, with the additional constraint $\tr(P^T \phi(X) D \phi(X)^T P)=1$, we can formulate the new objective function as follows:
\begin{equation}\label{objkmp}
\begin{split}
   & \argmin_{P, \alpha} \tr(P^T K L K P) \\
   & \text{s.t.}~ \tr(P^T K D K P)=1,~ \sum^{M}_{i=1} \alpha_i=1,~ \alpha_i \geq 0,
\end{split}
\end{equation}
or in the form associated with the norm constraint:
\begin{equation}\label{objkmpa}
\begin{split}
   & \argmin_{P, \alpha} \frac{\tr(P^T K L K P)}{\tr(P^T K D K P)}, ~ \text{s.t.}~ \sum^{M}_{i=1} \alpha_i=1,~ \alpha_i \geq 0.
\end{split}
\end{equation}

\subsection{Alternate Optimization via Relaxation}
In this section, we employ a procedure of alternate optimization \cite{DBLP:conf/afss/BezdekH02} to derive the solution of the optimization problem. To the best of our knowledge, it is difficult to find its optimal solution directly, especially for the weights in (\ref{objkmpa}).

First, for a fixed $\alpha$, finding the optimal projection $P$ is simply reduced to solve the generalized eigenvalue problem
\begin{equation}\label{kmpeigen}
  KLK \mathbf{p} = \lambda KDK \mathbf{p},
\end{equation}
and set $P = [\mathbf{p}_1, \cdots, \mathbf{p}_d]$ corresponds to the smallest $d$ eigenvalues based on the Ky-Fan theorem \cite{bhatia1997matrix}.

Next, to optimize $\alpha$, we derive a relaxed objective function from the original problem. The output of the relaxed function can ensure that the value of the objective function in (\ref{objkmpa}) is in a small neighborhood of the true minimum.

\begin{figure*}
\begin{align}\label{relax}
   \nonumber \frac{\tr(P^T K L K P)}{\tr(P^T K D K P)} & = \frac{\tr\Big(P^T (\alpha_1 K_1 + \alpha_2 K_2) (\alpha_1 L_1 + \alpha_2 L_2) (\alpha_1 K_1 + \alpha_2 K_2) P\Big)}{\tr\Big(P^T (\alpha_1 K_1 + \alpha_2 K_2) (\alpha_1 L_1 + \alpha_2 L_2) (\alpha_1 K_1 + \alpha_2 K_2) P\Big)} \\
     \nonumber & = \frac{\alpha_1^3 L_{111} + 2\alpha_1^2 \alpha_2 L_{121} + \alpha_1 \alpha_2^2 L_{221} + \alpha_1^2 \alpha_2 L_{112} + 2 \alpha_1 \alpha_2^2 L_{122} + \alpha_2^3 L_{222}}{\alpha_1^3 D_{111} + 2\alpha_1^2 \alpha_2 D_{121} + \alpha_1 \alpha_2^2 D_{221} + \alpha_1^2 \alpha_2 D_{112} + 2 \alpha_1 \alpha_2^2 D_{122} + \alpha_2^3 D_{222}}\\
     \nonumber & \leq \frac{1}{\alpha_1^3 L_{111} + 2\alpha_1^2 \alpha_2 L_{121} + \alpha_1 \alpha_2^2 L_{221} + \alpha_1^2 \alpha_2 L_{112} + 2 \alpha_1 \alpha_2^2 L_{122} + \alpha_2^3 L_{222}} \\
    \nonumber  & \quad \times \left(\frac{(\alpha_1^3 L_{111})^2}{\alpha_1^3 D_{111}} + \frac{(2\alpha_1^2 \alpha_2 L_{121})^2}{2\alpha_1^2 \alpha_2 D_{121}} + \frac{(\alpha_1 \alpha_2^2 L_{221})^2}{\alpha_1 \alpha_2^2 D_{221}} + \frac{(\alpha_1^2 \alpha_2 L_{112})^2}{\alpha_1^2 \alpha_2 D_{112}} + \frac{(2 \alpha_1 \alpha_2^2 L_{122})^2}{2 \alpha_1 \alpha_2^2 D_{122}} + \frac{(\alpha_2^3 L_{222})^2}{\alpha_2^3 D_{222}}\right) \\
     \nonumber & = \frac{1}{\alpha_1^3 L_{111} + 2\alpha_1^2 \alpha_2 L_{121} + \alpha_1 \alpha_2^2 L_{221} + \alpha_1^2 \alpha_2 L_{112} + 2 \alpha_1 \alpha_2^2 L_{122} + \alpha_2^3 L_{222}} \times \Big(\alpha_1^3 L_{111} \frac{L_{111}}{D_{111}} \\
     \nonumber & \quad + 2\alpha_1^2 \alpha_2 L_{121} \frac{L_{121}}{D_{121}} + \alpha_1 \alpha_2^2 L_{221} \frac{L_{221}}{D_{221}} + \alpha_1^2 \alpha_2 L_{112} \frac{L_{112}}{D_{112}} + 2 \alpha_1 \alpha_2^2 L_{122} \frac{L_{122}}{D_{122}} + \alpha_2^3 L_{222} \frac{L_{222}}{D_{222}}\Big)\\
      & = \sum_{i,j,k \in \{1,2\}} w_{ijk}(\alpha_1, \alpha_2) \frac{L_{ijk}}{D_{ijk}}.
\end{align}
\end{figure*}

We fix the projection $P$ to update $\alpha$ individually. Without loss of generality, we first consider the condition that $M = 2$, i.e., there are only two views. Then the optimization problem (\ref{objkmpa}) is reduced to
\begin{equation}\label{objkmp2}
\begin{split}
   & \argmin_{P, \alpha} \frac{\tr(P^T K L K P)}{\tr(P^T K D K P)},~ \alpha_1 + \alpha_2 =1,~ \alpha_1, \alpha_2 \geq 0.
\end{split}
\end{equation}
For simplicity, we denote $L_{ijk} = \tr(P^T K_i L_k K_j P)$ and $D_{ijk} = \tr(P^T K_i D_k K_j P)$, $i, j, k \in \{1, 2\}$. Then we can simply find that $L_{ijk} = L_{jik}$ and $D_{ijk} = D_{jik}$.

\paragraph{Relaxation} With the Cauchy-Schwarz inequality \cite{hardy1952inequalities}, the relaxation for the objective function in (\ref{objkmp2}) is shown in Eq. (\ref{relax}), where $w_{ijk}$ is the coefficient of $\frac{L_{ijk}}{D_{ijk}}$ and $\sum_{i,j,k \in \{1,2\}} w_{ijk} =1$.
In this way, the objective function in (\ref{objkmp2}) is relaxed to a weighted sum of $\frac{L_{ijk}}{D_{ijk}}$. Thus, minimizing the weighted sum of the right-hand-side in (\ref{relax})
can lower the objective function value in (\ref{objkmp2}). Note that
\begin{equation*}
\alpha_1^2 \alpha_1 = \frac{1}{2} \alpha_1 \cdot \alpha_1 \cdot 2\alpha_2 \leq \frac{1}{2}\left(\frac{\alpha_1 + \alpha_1 + 2\alpha_2}{3}\right)^3 = \frac{4}{27},
\end{equation*}
and then the weights without containing $\alpha_1^3$ and $\alpha_2^3$ are always smaller than a constant. Therefore, we only ensure that a part of the terms in the weighted sum is minimized, i.e., to solve the following optimization problem:
\begin{equation}\label{kmprel2w}
  \argmin_{\alpha_1, \alpha_2} w_{111} \frac{L_{111}}{D_{111}} + w_{222} \frac{L_{222}}{D_{222}}, ~\text{s.t.}~ w_{111} + w_{222} = 1.
\end{equation}
Since $w_{111}$ and $w_{222}$ are the functions of $(\alpha_1, \alpha_2)$, we first find the optimal weights without parameters $(\alpha_1, \alpha_2)$. To avoid trivial solution, we assign an exponent $r > 1$ for each weight. By denoting $\gamma_1 = w_{111}$ and $\gamma_2 = w_{222}$, the relaxed optimization will be
\begin{equation}\label{kmprel2}
  \argmin_{\gamma_1, \gamma_2} \gamma_1^r \frac{L_{111}}{D_{111}} + \gamma_2^r \frac{L_{222}}{D_{222}}, ~\text{s.t.}~ \gamma_1 + \gamma_2 =1, \gamma_1, \gamma_2 \geq 0.
\end{equation}

For (\ref{kmprel2}), we have the Lagrangian function with the Lagrangian multiplier $\eta$:
\begin{equation}
  L(\gamma_1,\gamma_2,\eta) = \gamma_1^r \frac{L_{111}}{D_{111}} + \gamma_2^r \frac{L_{222}}{D_{222}} - \eta (\gamma_1 + \gamma_2 -1 ).
\end{equation}
We only need to set the derivatives of $L$ with respect to $\gamma_1$, $\gamma_2$ and $\eta$ to zeros as follows:
\begin{eqnarray}
  \frac{\partial L}{\partial \gamma_1} &=& r \gamma_1^{r-1} \frac{L_{111}}{D_{111}} - \eta =0, \\
  \frac{\partial L}{\partial \gamma_2} &=& r \gamma_2^{r-1} \frac{L_{222}}{D_{222}} - \eta =0,  \\
  \frac{\partial L}{\partial \eta} &=& \gamma_1 + \gamma_2 -1 =0.
\end{eqnarray}
Then $\gamma_1$ and $\gamma_2$ can be calculated by
\begin{equation}
\begin{split}
  \gamma_1 & = \frac{(L_{222} D_{111})^{\frac{1}{r-1}}}{(L_{222} D_{111})^{\frac{1}{r-1}} + (L_{111} D_{222})^{\frac{1}{r-1}}}, \\
   \gamma_2 & = \frac{(L_{111} D_{222})^{\frac{1}{r-1}}}{(L_{222} D_{111})^{\frac{1}{r-1}} + (L_{111} D_{222})^{\frac{1}{r-1}}}.
\end{split}
\end{equation}

Having acquired $\gamma_1$ and $\gamma_2$, we can obtain $\alpha_1$ and $\alpha_2$ by the corresponding relationship between the coefficients of the functions in (\ref{kmprel2w}) and (\ref{kmprel2}):
\begin{equation}
  \frac{\alpha_1^3 L_{111}}{\alpha_2^3 L_{222}} = \frac{w_{111}}{w_{222}} = \frac{\gamma_1^r}{\gamma_2^r}.
\end{equation}
With the constraint $\alpha_1 + \alpha_2 =1$, we can easily find that
\begin{equation}
  \begin{split}
    \alpha_1 & = \frac{(\gamma_1^r L_{222})^{\frac{1}{3}}}{(\gamma_1^r L_{222})^{\frac{1}{3}} + (\gamma_2^r L_{111})^{\frac{1}{3}}}, \\
    \alpha_2 & = \frac{(\gamma_2^r L_{111})^{\frac{1}{3}}}{(\gamma_1^r L_{222})^{\frac{1}{3}} + (\gamma_2^r L_{111})^{\frac{1}{3}}}.
  \end{split}
\end{equation}

Hence, for the general $M$-view situation, we also have the corresponding relaxed problems:
\begin{equation}\label{kmprelaxw}
  \argmin_{\sum^M_{i=1} \alpha_i =1} \sum_{i,j,k \in \{1, \cdots, M\}} w_{ijk}(\alpha_1, \cdots, \alpha_M) \frac{L_{ijk}}{D_{ijk}}
\end{equation}
and
\begin{equation}\label{kmprelaxm}
  \argmin_{\gamma_1, \cdots, \gamma_M} \sum^M_{i=1} \gamma_i^r \frac{L_{iii}}{D_{iii}}, ~\text{s.t.}~ \sum^M_{i=1} \gamma_i = 1,~ \gamma_i \geq 0.
\end{equation}
The coefficients $(\gamma_1, \cdots, \gamma_M)$ and $(\alpha_1, \cdots, \alpha_M)$ can be obtained in similar forms:
\begin{equation}\label{beta}
  \gamma_i = \frac{(D_{iii}/L_{iii})^{\frac{1}{r-1}}}{\sum^M_{j=1}(D_{jjj}/L_{jjj})^{\frac{1}{r-1}}},~ i=1,\cdots, M
\end{equation}
and
\begin{equation}\label{alpha}
  \alpha_i = \frac{(\gamma_i^r/L_{iii})^{\frac{1}{3}}}{\sum^M_{j=1} (\gamma_j^r/L_{jjj})^{\frac{1}{3}}},~ i=1,\cdots, M.
\end{equation}

\paragraph{Convergence} Although the weight $\alpha$ obtained in the above procedure is not the global minimum, the objective function is ensured in a range of small values. We let $F_1$ and $F_2$ be the objective functions in (\ref{objkmpa}) and (\ref{kmprelaxw}), respectively, and let
\begin{equation}
  F_3 = \sum_{i=j=k} w_{ijk} \frac{L_{ijk}}{D_{ijk}} = \sum^M_{i=1} w_{iii} \frac{L_{iii}}{D_{iii}}.
\end{equation}
We can find that $F_1 \leq F_2$ and if there exists $\alpha_i = 1$ for some $i$, then $F_1 = F_2 = F_3$. During the alternate procedure, for optimizing $P$, $F_1$ is minimized, and for optimizing $\alpha$, $F_3$ is minimized. Denote $m_1 = \max (F_1 - F_3)$ and $(P_1, \alpha_1) = \argmax (F_1 - F_3)$, then we have
\begin{equation*}
\begin{split}
\min F_3 + m_1 &\leq F_3 (P_1, \alpha_1) + (F_1 - F_3)(P_1, \alpha_1) \\
    & = F_1 (P_1, \alpha_1) \leq \max F_1,
\end{split}
\end{equation*}
and we can define the following nonnegative continuous function:
\begin{equation}\label{f4}
  F_4 (P,\alpha) = \max\Big(F_1(P, \alpha), \min_{\alpha} \big(F_3(P, \alpha) + m_1\big)\Big).
\end{equation}

Note that $\min_{\alpha} \big(F_3(P, \alpha) + m_1\big)$ is independent of $\alpha$, thus for any $P$, there exists $\alpha_0$, such that $F_1 (P, \alpha_0) = \min_{\alpha} \big(F_3(P, \alpha) + m_1\big)$. If we impose the above alternate optimization on $F_4$, $F_4$ is nonincreasing and therefore converges. Though $\alpha$ does not converge to a fixed point, the value of $F_1$ is reduced into a small district, which is smaller than $\min_{\alpha} F_3$ plus a constant. It is also worthwhile to note that $F_3$ is actually the weighted sum of the objective functions for preserving each view's locality information. However, the optimization for $F_3$ still learns information from each view separately, i.e., the locality similarity is not fused. We summarize the KMP in Algorithm \ref{al1}.

\begin{algorithm}
\caption{Kernelized Multiview Projection}
\label{al1}
\begin{algorithmic}[1]
\REQUIRE
      The training samples $\{S_1, \cdots, S_N\}$ and parameter $r>1$.
\ENSURE
      The projection matrix $P \in \mathbb{R}^{N \times d}$ and the weights $\alpha = (\alpha_1, \cdots, \alpha_M) \in \mathbb{R}^{M}$ for kernel matrices.
\STATE Extract multiple features from each training image and obtain data matrices $X^i_p$, $p=1, \cdots, N,~ i= 1, \cdots, M$;\STATE Compute the similarity matrices $W_1, \cdots, W_M$ and the Laplacian matrices $L_1 \cdots, L_M$ by solving the optimization problem in (\ref{l1graph}) for each view;
\STATE Compute the kernel matrices $K_1, \cdots, K_M \in \mathbb{R}^{N \times N}$ and the Laplacian matrices $L_1, \cdots, L_M \in \mathbb{R}^{N \times N}$ for $M$ views;
\STATE Initialize $\alpha \leftarrow (\frac{1}{M}, \cdots, \frac{1}{M})$;
\REPEAT
    \STATE Compute the fused kernel matrix $K = \sum^M_{i=1} \alpha_i K_i$ and the fused Laplacian matrix $L = \sum^M_{i=1} \alpha_i L_i$;
    \STATE Compute $P$ by solving the generalized eigenvalue problem (\ref{kmpeigen});
    \STATE Compute coefficients $\gamma = (\gamma_1, \cdots, \gamma_M)$ by Eq. (\ref{beta});
    \STATE Transform $\gamma$ to $\alpha$ by Eq. (\ref{alpha});
\UNTIL $F_4$ defined in Eq. (\ref{f4}) converges.
\end{algorithmic}
\end{algorithm}

\section{Experiments and Results}\label{exp}

In this section, we evaluate our Kernelized Multiview Projection (KMP) on three image datasets: CMU PIE, CIFAR10 and SUN397 respectively. The \textbf{CMU PIE} face dataset \cite{DBLP:journals/vldb/CaiHH11} contains $41,368$ images from 68 subjects (people). Following the settings in \cite{DBLP:journals/vldb/CaiHH11}, we select $11,554$ front face images, which are manually aligned and cropped into $32\times32$ pixels. Further, $7,500$ images are used as the training set and the remaining $4,054$ images are used for testing. The \textbf{CIFAR10} dataset \cite{DBLP:journals/pami/TorralbaFF08} is a labeled subset of the $80$-million tiny images collection. It consists of a total of $60,000$ $32\times32$ color images in $10$ classes. The entire dataset is partitioned into two parts: a training set with $50,000$ samples and a test set with $10,000$ samples. The \textbf{SUN397} dataset \cite{DBLP:conf/cvpr/XiaoHEOT10} contains $108,754$ scene images in total from $397$ well-sampled categories with at least $100$ images per category. We randomly select $50$ samples from each category to construct the training set and the rest of samples are the test set. Thus, there are $19,850$ and $88,904$ images in the training set and test set, respectively.

\begin{table}
\caption{Dimensions of four features for image classification.}
\begin{center}
\newcommand{\tabincell}[2]{\begin{tabular}{@{}#1@{}}#2\end{tabular}}
\label{table:T1}
\begin{tabular}{|c|c|}
\hline 
 \textbf{Feature representation} & \textbf{Dimension}\\
 \hline\hline
Histogram of oriented gradients (HOG)&225\\
\hline
Local binary pattern (LBP)&256\\
\hline
Color histogram (ColorHist)&192\\
\hline
GIST&384\\
\hline
\textbf{Total dimension}&1057\\
\hline
\end{tabular}
\end{center}
\vspace{-1ex}
\end{table}

\subsection{Compared Methods and Settings}
For image classification, each image can be usually described by different feature representations, i.e., multiview representation, in high-dimensional feature spaces. In this paper, we adopt four different feature representations: HOG \cite{dalal2005histograms}, LBP \cite{ahonen2004face}, ColorHist and GIST \cite{DBLP:journals/ijcv/OlivaT01} to describe each image. Table~\ref{table:T1} illustrates the original dimensions of these features.

\begin{table}
\small
\caption{Performance comparison (\%) between the SVM using multiple features through KMP and the SVM using single original features. The numbers in parentheses indicate the dimensions of the representations. For MKL-SVM, $\ell^{1}$-graph is also used to construct the kernel matrix for each view and then MKL-SVM is applied to final classification.}
\label{table:T2}
\begin{center}
\begin{tabular}{|c|c|c|c|}
\hline
\backslashbox{\textbf{Method}}{\textbf{Dataset}}& \textbf{CMU PIE}&\textbf{CIFAR10}&\textbf{SUN397}\\
\hline\hline
HOG&83.3& 70.2& 29.3\\
LBP&74.6& 54.2&20.4\\
ColorHist&31.2&23.0&9.3\\
GIST&94.2&82.3&17.5\\
\hline
Concatenation& 93.4& 82.8&31.9\\
\hline
MKL-SVM & 95.6& 86.3& 30.7\\
\hline
KMP&\textbf{99.5}(60) & \textbf{89.7}(80)&\textbf{40.5}(70)\\
\hline
\end{tabular}
\end{center}
\vspace{-1ex}
\end{table}

We compare our proposed KMP with two related multi-kernel fusion methods. In particular, the RBF kernels\footnote{Our approach can work with any legitimate kernel function, though we focus on the popular RBF kernel in this paper} for each view are adopted in the proposed KMP method: $$K = \sum_{i=1}^M \alpha_i K_i,$$ where the weight $\alpha_{i}$ is obtained via alternate optimization. AM indicates that the kernels are combined by arithmetic mean: $$K_{AM} =\frac{1}{M}\sum_{i=1}^M K_i,$$ and GM denotes the combination of kernels through geometric mean: $$K_{GM} =(\prod_{i=1}^M K_i)^{\frac{1}{M}}.$$ Besides, we also include the best performance of the single-view-based spectral projection (BSP), the average performance of the single-view-based spectral projection (ASP) and the concatenation of single-view-based embeddings (CSP) in our compared experiments. In particular, AM and GM are incorporated with the proposed KMP framework. BSP, ASP and CSP are based on the kernelized extension of Discriminative Partition Sparsity Analysis (DPSA) \cite{DBLP:conf/icpr/LiuS14} technique. In addition, two non-linear embedding methods, distributed spectral embedding (DSE) and multiview spectral embedding (MSE), are adopted in our comparison, as well. In DSE and MSE, the Laplacian eigenmap (LE) \cite{belkin2001laplacian} is adopted. For all these compared embedding methods, the RBF-SVM is adopted to evaluate the final performance.

All of the above methods are then evaluated on seven different lengths of codes: $\{20, 30, 40, 50, 60, 70, 80\}$. Under the same experimental setting, all the parameters used in the compared methods have been strictly chosen according to their original papers. For KMP and MSE, the optimal balance parameter $r$ for each dataset is selected from one of $\{2, 3, 4, 5, 6, 7, 8, 9, 10\}$, which yields the best performance by 10-fold cross-validation on the training set. The number of the GMM clusters $G$ in KMP is selected from one of $\{10, 20,\ldots,100\}$ with a step of 10 via cross-validation on the training data. The same procedure occurs on the selection of sparsity hyperparameter $\varepsilon$ from one of $\{5,8,10,12,15,18,20\}$. The best smooth parameter $\sigma$ in the construction of the RBF kernel and RBF-SVM is also chosen by the cross-validation on the training data. Since the clustering procedure has uncertainty, all experiments are performed five times repeatedly and each of the results in the following section is the averages of five runs.
\begin{figure*}
  \centering
  \includegraphics[width=0.95\textwidth]{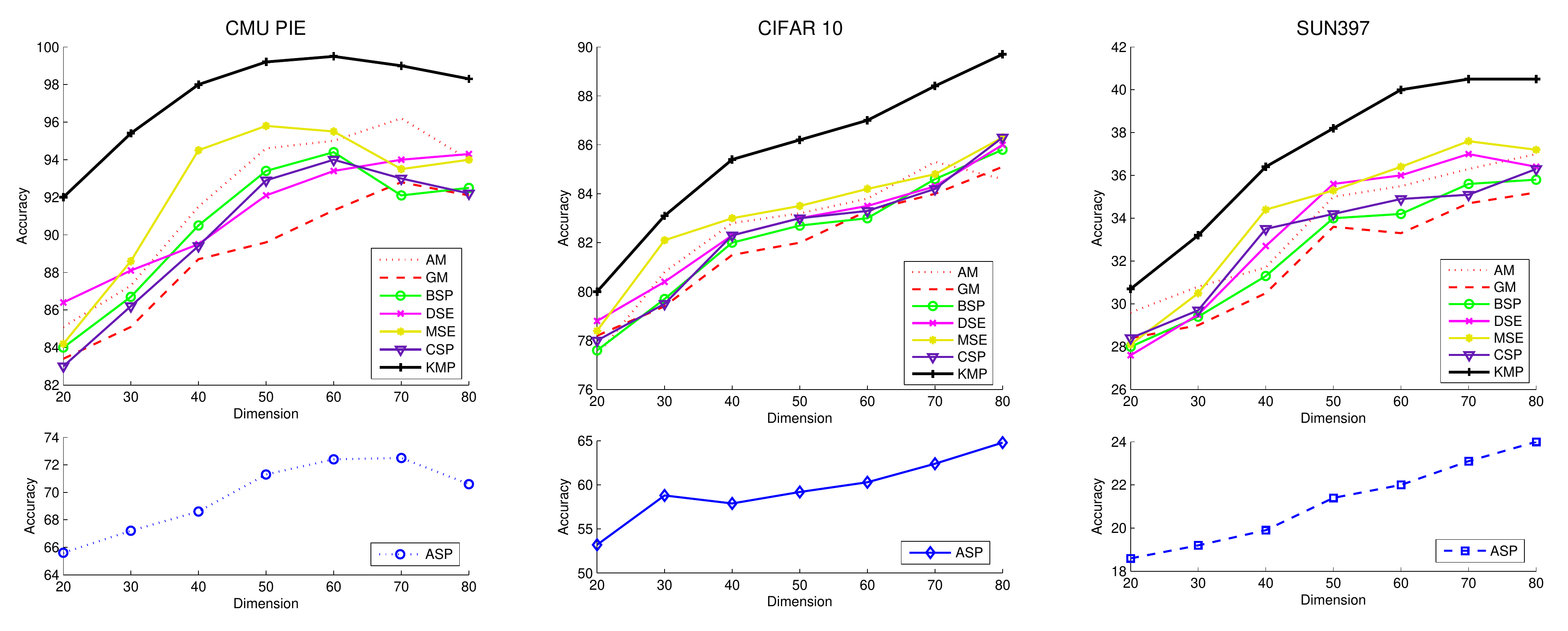}\\
  \caption{Performance comparison (\%) of KMP with different multiview embedding methods on the three datasets.}\label{acc}
\end{figure*}

\subsection{Results}
In Table~\ref{table:T2}, we first illustrate the performance of the original single-view representations on all the three datasets. In detail, we extract original feature representations under one certain view and then directly feed them to the SVM for classification. From the comparison, we can easily observe that the GIST features consistently outperform the other descriptors on the CMU PIE and CIFAR10 datasets but HOG takes the superior place on the SUN397 dataset. The lowest accuracy is always obtained by ColorHist. Furthermore, we also include the long representation, which is concatenated by all the four original feature representations, into this comparison. It is shown that in most of the time the concatenated representation can reach better performance than single view representations, but is always significantly worse than the proposed KMP. Additionally, the results of the multiple kernel learning based on SVM (MKL-SVM) \cite{DBLP:journals/jmlr/GonenA11} are listed in Table~\ref{table:T2} using the same four feature descriptors. Specifically, the best accuracies achieved by KMP are 99.5\%, 89.7\% and 40.5\% on the CMU PIE, CIFAR10, and SUN397, respectively.

\begin{figure*}
\centering
\includegraphics[width=0.95\textwidth]{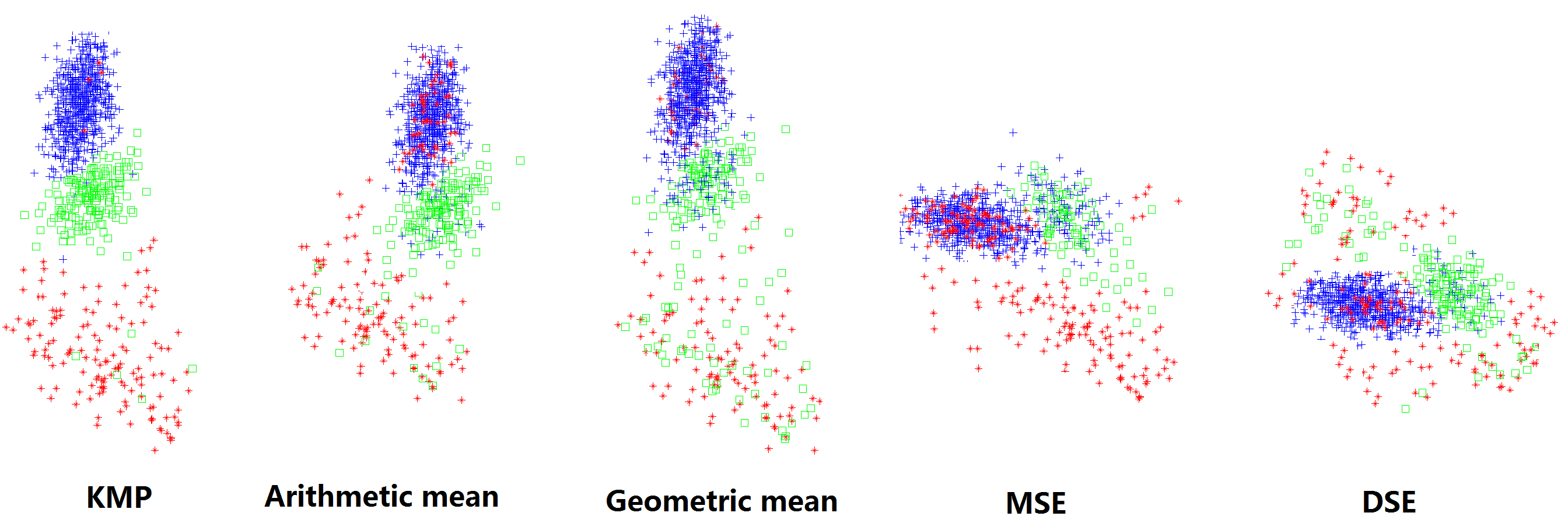}
\caption{Illustration of low-dimensional distributions of five different fusion schemes (illustrated with data of three categories from the CIFAR10 dataset).}
\label{fig:Pe}
\end{figure*}

In Fig. \ref{acc}, seven different embedding schemes are compared with the proposed KMP on all the three datasets. From the comparison, the proposed KMP always leads to the best performance for image classification. Meanwhile, arithmetic mean (AM) and the single-view-based spectral projection (BSP) generally achieve higher accuracies than the best performance of geometric mean (GM) and the average performance of the single-view-based spectral projection (ASP). The concatenation of single-view-based embeddings (CSP) achieves competitive performance compared with BSP on all the three datasets. DSE always produces worse performance than MSE and sometimes even obtains lower results than CSP. However, DSE generates better performance than GM and ASP, since a more meaningful multiview combination scheme is adopted in DSE. Beyond that, it is obviously observed that, with different target dimensions, there are large differences among the final results. Fig.~\ref{fig:Pe} plots the low-dimensional embedding results obtained by AM, GM, KMP, DSE and MSE on the CIFAR10 dataset. Our proposed KMP can well separate different categories, since it takes the semantically meaningful data structure of different views into consideration for embedding.

\begin{table}
\tiny
\caption{Performance (\%) of KMP with different $r$ values on the CMU PIE dataset.}
\centering
\label{table:T6}
\begin{center}
\begin{tabular}{|c|c|c|c|c|c|c|c|c|c|}
\hline
Dimension& r=2 & r=3 & r=4&r=5&r=6&r=7&r=8&r=9&r=10\\
\hline\hline
d=20& 87.0& 87.0& 87.5& 87.8&\textbf{88.9}& 88.7& 88.0&88.0&87.4\\
d=30& 89.4& 90.1&  90.5& 91.0&91.3&\textbf{91.4}& \textbf{91.4}&90.7&89.3\\
d=40& 87.2& 89.0& 89.4& 91.2& 92.0& 93.5& 93.5&\textbf{93.7}&93.2\\
d=50& 84.8& 95.1& 95.5& 96.0& 96.4& 97.3& \textbf{98.2}&97.9&97.5\\
d=60& 97.3& 97.5& 98.4& 98.2& 98.7& 99.2& 99.6&\textbf{99.8}&99.7\\
d=70& 96.2& 96.4& 96.9& 97.2& 97.9& 98.2& 98.5&\textbf{99.0}&98.7\\
d=80& 96.5& 96.8& 97.2& 97.5& 97.1& 97.4& 98.0&98.3&\textbf{98.6}\\
\hline
\end{tabular}
\end{center}
\vspace{-2ex}
\end{table}

In addition, we can observe that with the increase of the dimension, all the curves of compared methods on the CIFAR10 and SUN397 datasets are climbing up except for DSE and MSE, both of which have a slight decrease on SUN397 when the dimension exceeds $70$. However, on the CMU PIE dataset, the results in comparison always climb up then go down for almost every compared method except for DSE when the length of dimension increases (see Fig.~\ref{acc}). For instance, the highest accuracy on the CMU PIE dataset is on the dimension of $60$ and the best performance on CIFAR10 and SUN397 happens when $d=80$ and $d=70$, respectively.

\begin{figure}
  \centering
  \includegraphics[width=0.49\textwidth]{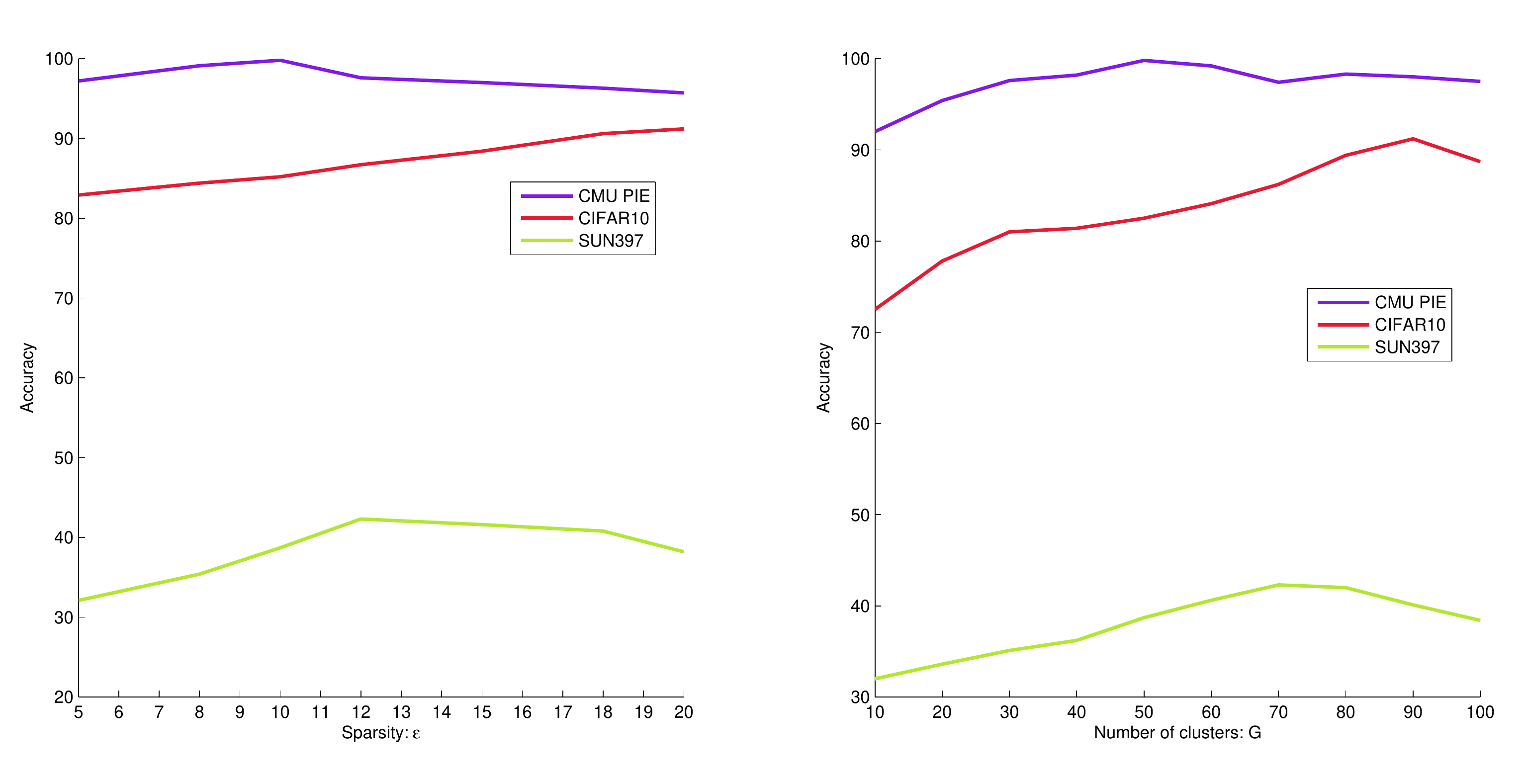}\\
  \caption{The curves on the left side show the best performance on the training data when $\varepsilon$ is equal to one value from $\{5,8,10,12,15,18,20\}$ while $G$ varies its value in $\{10, 20, \cdots, 100\}$, and vice versa.}\label{param}
\end{figure}

Furthermore, some parameter sensitivity analysis is carried out. Table~\ref{table:T6} illustrates the performance variation of KMP with respect to the parameter $r$ on the CMU PIE dataset; the target dimensionality of the low-dimensional embedding $d$ is fixed at $\{20, 30,\ldots, 80\}$ with a step of 10, respectively. By adopting the 10-fold cross-validation scheme on the training data, it is demonstrated that higher dimensions prefer a larger $r$ in our KMP. Finally, Fig.~\ref{param} shows the variation of parameters $G$ and $\varepsilon$ on all three datasets. The general tendency of these curves is consistently shown as ``rise-then-fall''. It can be also seen from this figure that a larger training set needs larger values of $G$ and $\varepsilon$, and vice versa.

\begin{table}
\begin{center}
\newcommand{\tabincell}[2]{\begin{tabular}{@{}#1@{}}#2\end{tabular}}
\scriptsize
\caption{Comparison of training and coding time (seconds) for learning 80 dimensional embedded features on the three datasets.}
\label{time}
\begin{tabular}{|c|c|c|c|c|c|c|c|}
\hline
\textbf{Dataset}&\textbf{Phase}&DSE & MSE &MKL&\textbf{KMP}\\
\hline\hline
\textbf{CMU PIE}&Training time&1148.24 &716.79 &873.72 &755.28 \\
&Coding time/query&1156.01 &790.09 &- &0.032 \\
\hline\hline
\textbf{CIFAR 10}&Training time&1683.70 &1026.32 &1098.97 &991.54 \\
&Coding time/query&1696.52 &1072.18 & -& 0.041\\
\hline\hline
\textbf{SUN397}&Training time&2804.91 &1778.74 &1678.14 &1694.10 \\
&Coding time/query&2812.36 &1784.50 &- &0.036 \\
\hline
\end{tabular}
\end{center}
\vspace{-2ex}
\end{table}

\subsection{Time Consumption Analysis}
In this section, we compare the training and coding time of the proposed KMP algorithm with other methods. As we can see from Table \ref{time}, our method can achieve competitive training time compared with the state-of-the-art multiview and multiple kernel learning methods. Since there is no embedding procedure in MKL, the coding time is not applicable for MKL. Due to the nature of DSE and MSE, they need to be re-trained when receiving a new test sample. In contrast, once the projection and weights are gained by KMP, they are fixed for all test samples and implemented in a fast way. All the experiments are completed using Matlab 2014a on a workstation configured with an i7 processor and 32GB RAM.

\section{Conclusion}\label{con}
In this paper, we have presented an effective subspace learning framework called Kernelized Multiview Projection (KMP). KMP, as an unsupervised method, can encode a variety of features in different ways, to achieve a semantically meaningful embedding. Specifically, KMP is able to successfully explore the complementary property of different views and finally find the low-dimensional subspace where the distribution of each view is sufficiently smooth and discriminative. KMP can be regarded as a fused dimensionality reduction method for multiview data. We have objectively evaluated our approach on three datasets: CMU PIE, CIFAR10 and SUN397. The corresponding results have shown the effectiveness and the superiority of our algorithm compared with other multiview embedding methods. For future work, we plan to combine the current KMP approach with semi-supervised learning for other computer vision tasks.

{\small
\bibliographystyle{ieee}
\bibliography{KMP}
}

\end{document}